\documentclass{bmvc2k}

\usepackage{amssymb,amsthm,mathtools}
\usepackage{booktabs}
\usepackage{multirow}
\usepackage{microtype}
\usepackage{enumitem}
\usepackage{tikz}
\usepackage{listings}
\usepackage{tabularx}
\usetikzlibrary{arrows.meta,positioning,fit,calc}

\newcommand{\E}{\mathbb{E}}

\newcommand{\Normal}{\mathcal{N}}
\newcommand{\eps}{\epsilon}

\newcommand{\norm}[1]{\left\lVert #1\right\rVert}

\title{RAIN: Region-Aware Inversion Network for Semantic Watermark Extraction}
\addauthor{Zilai Li}{}{1}
\addinstitution{Independent Researcher}
\runninghead{Li}{RAIN: Region-Aware Inversion Network}

\begin{document}
\maketitle

\begin{abstract}
Semantic watermarks for diffusion models embed ownership information into the generative process while preserving perceptual quality, but the watermark extraction process in some algorithms, like Gaussian Shading, conventionally requires multi-step diffusion inversion to recover the initial noise. Recent one-step methods show that this cost can be reduced substantially. We study this problem through extended rectified flow and conditional regression. The key observation is that, near the high-SNR image endpoint, recovering a useful noise statistic given by the first-step output of the extended rectified flow is much simpler than reconstructing the full inverse trajectory, and that is enough to extract the watermark, since extracting a watermark given by Gaussian Shading only requires knowing which Gaussian region the original latent occupies. Based on this observation, we propose a lightweight, prompt-free extractor to predict the noise statistics. Meanwhile, we further decompose endpoint statistic recovery into an image-like anchor and a noise-oriented residual, thereby increasing the model's ability to leverage GPU parallel computation. The resulting method avoids iterative inversion and repeated evaluation of a diffusion-scale U-Net, providing an efficient one-step extraction pipeline with a concise theoretical interpretation. For the Gaussian Shading watermark extraction process, the computational cost of our algorithm is 14.394 GFLOPs, which is 47 times fewer than FARI and OSI.
\end{abstract}

\section{Introduction}

Semantic watermarks for diffusion models \cite{yang2024gaussianshading,wen2023tree,huang2024robin,ci2024ringid, gunn2025undetectable,yang2026t2smark} embed ownership information into the generative process rather than modifying the final image after generation. Gaussian Shading (GS)~\citep{yang2024gaussianshading} is a representative example: it encodes a secret message into the initial Gaussian noise while preserving its marginal Gaussian distribution. However, conventional GS watermark extraction requires multi-step DDIM inversion \cite{mokady2023null} to recover the noise from the generated image, and reducing the computation cost of DDIM inverse for watermark extraction is important.  Although there exist algorithms aiming to improve DDIM inversion \cite{wang2024belm,wallace2023edict}, or reducing inference cost \cite{zhou2024amed, tong2025learning, chen2024trajectory, yin2024dmd, yin2024improveddmd, luo2311lcm, song2023consistency}, neither of these algorithms is for watermark extraction. Recent one-step watermark extraction methods \cite{chen2026osi,faricode2026} suggest that this expensive inverse process may be unnecessary.

Our first observation is that regression does not, in general, recover the particular noise that generated an observed image. If several noises are the starting points of the inference trajectory that ends at the same or similar image, an $L_2$ regression objective learns the mean of the noise.  And this situation is possible, since the Gaussian noise is in a vast space, while the image only occupies a low-dimensional manifold. This is closely related to the intuition behind Tweedie's formula \cite{efron2011tweedie}: under Gaussian corruption, when we sample a noisy image without knowing the original image, the best prediction, in the sense of minimizing the L2 distance between the prediction and the true image, is the mean over the possible images that can generate this noisy image via adding random noise. For Gaussian Shading, knowing the mean can be sufficient because watermark extraction only requires the predicted noise to preserve the encoded Gaussian regions.  If different noises generate the same image, then DDIM inverse will produce a noise sample different from the original noise, but invalid watermark extraction is rare in normal DDIM inverse process. So we further assume that predicting the mean or median of these noises is also sufficient and does not invalidate the GS, which motivates us to propose an algorithm aiming to model the statistics of noise in a particular region to extract the corresponding watermark: region-aware inverse network.  

Our second observation comes from Rectified Flow~\citep{liu2023rectifiedflow}: In the high SNR region, the RF-2 algorithm will study the velocity field point to the means of noise that generate the current noisy image. Explicitly, the interpolation between image and noise to get the noisy image in the training is non-causal, and that cannot be used in inference; the regression loss lets the model learn the conditional mean velocity of the possible trajectories passing through the current noisy image. Iteratively following this causal velocity field can still reproduce the desired image distribution. Therefore, predicting a conditional mean rather than a particular paired endpoint is already a standard mechanism of generative flow models.

This perspective makes one-step watermark extraction relate to an especially simple task that generates an image in the high-SNR regime of inference. Explicitly, for the RF-2 algorithm, in the high-SNR regime, most of the generation process has already been completed, and the network only needs to learn the mean of the velocity that can complete the remaining part of the transport, which is an easy task, and it's equivalent to predicting the mean of the possible noise to generate the final image, a useful statistic for watermark extraction. We therefore hypothesize that this high-SNR watermark extraction task can be handled by a substantially smaller network than a full diffusion denoiser.

Based on these observations, we propose an aggressive lightweight estimator that directly predicts recoverable noise statistics from an almost-clean image. Our method avoids iterative inversion and decomposes the endpoint regression into easier components, substantially reducing the computational cost of watermark extraction.  The inference computation and parameter count are both lower than the latest extraction methods, such as OSI and FARI.

\section{Background}
\subsection{From DDPM Corruption to Extended Flow Matching and Rectified Flow}

The DDPM algorithm use a special diffusion process~\citep{ho2020ddpm} to gradually perturbs the sample image via random noise through a Markov chain,
\begin{equation}
    q(X_t\mid X_{t-1})
    =
    \Normal\!\left(
        \sqrt{1-\beta_t}\,X_{t-1},
        \beta_t I
    \right),
\end{equation}
where $\{\beta_t\}$ is a predefined noise schedule. 
Although this process adds noise step by step, the renowned property of Gaussian transitions \cite{ho2020ddpm} allows us to sample the noisy state at any timestep directly:
\begin{equation}
    X_t
    =
    \sqrt{\bar\alpha_t}\,X_0
    +
    \sqrt{1-\bar\alpha_t}\,\eps,
    \qquad
    \eps\sim\Normal(0,I),
    \label{eq:ddpm_path}
\end{equation}
with $\bar \alpha_t=\prod_{i=1}^{t}(1-\beta_i)$.

Equation~\eqref{eq:ddpm_path} reveals a useful interpretation of diffusion training. 
Instead of viewing DDPM only as a sequence of stochastic perturbations, each noisy sample $X_t$ can be viewed as a point on a trajectory given by interpolation between a clean endpoint $X_0$ and a Gaussian-noise endpoint $\eps$. 
In the standard DDPM construction,
\begin{equation}
    X_0\sim p_{\rm data},
    \qquad
    \eps\sim\Normal(0,I),
    \qquad
    X_0\perp\eps,
\end{equation}
so the two endpoints are sampled independently.

This interpolation viewpoint naturally connects DDPM to Flow Matching and Rectified Flow. 
More generally, let
\begin{equation}
    (X_0,\eps)\sim\pi_{0,\eps}
\end{equation}
denote an \emph{arbitrary coupling} between an image endpoint and a noise endpoint, and consider
\begin{equation}
    X_t
    =
    a_t X_0+b_t\eps,
    \label{eq:extended_path}
\end{equation}
where $a_t$ and $b_t$ are differentiable schedules. In the standard rectifly flow, $a_t=1-t, b_t =t$, the $t$ is continuous and is between zero to one.
For a clean-to-noise convention, one may take
$a_0\approx1,b_0\approx0$ and
$a_1\approx0,b_1\approx1$.
The corresponding non-causal velocity is
\begin{equation}
    U_t
    =
    \dot{a}_t X_0
    +
    \dot{b}_t\eps.
    \label{eq:conditional_velocity}
\end{equation}
, in which $\dot{a}_t$ is the deviation of the $b_t$ with respect to $t$, and so the $\dot{b}_t$.

Once both endpoints are known, 
Eqs.~\eqref{eq:extended_path}--\eqref{eq:conditional_velocity}
define a valid trajectory.
However, this trajectory is \emph{non-causal as an inference rule}: 
$U_t$ generally depends on the hidden pair $(X_0,\eps)$ and therefore cannot be determined from the current state $X_t$ alone.
Consequently, the sample-wise interpolation is not yet a closed ODE that can be integrated during inference.

Flow matching resolves this problem by learning a deterministic vector field from the observable state. 
Under the squared regression objective
\begin{equation}
    \min_v
    \;
    \E\!\left[
        \norm{v(X_t,t)-U_t}_2^2
    \right],
    \label{eq:fm_loss}
\end{equation}
the population-optimal field~\citep{bertrand2025closed} is:
\begin{equation}
    v^*(x,t)
    =
    \E[U_t\mid X_t=x].
    \label{eq:fm_mean}
\end{equation}


Importantly, rectified flow does not require the noise and image to be independent. 
Standard DDPM is only one special case, obtained by choosing the diffusion schedule in Eq.~\eqref{eq:ddpm_path} together with the independent coupling
$X_0\perp\eps$.
More generally, $X_0$ and $\eps$ may be deterministically coupled, produced by a teacher sampler\cite{liu2022rectified}. 


\subsection{Gaussian Shading}

Gaussian Shading (GS)~\citep{yang2024gaussianshading} embeds a watermark directly into the terminal Gaussian latent while preserving its marginal distribution. 
Let the latent have shape $c\times h\times w$, let $l$ denote the number of embedded bits per scalar in the high-dimensional Gaussian noise, and let $f_c$ and $f_{hw}$ denote the channel and spatial repetition factors. 
The effective payload length is
\begin{equation}
    k=
    \left\lfloor
    \frac{lchw}{f_c f_{hw}^2}
    \right\rfloor bits.
\end{equation}
The payload is first repeated according to $f_c$ and $f_{hw}$ and then encrypted with a stream cipher, producing an approximately uniform pseudorandom bit stream.

GS uses the encrypted bits to select equal-probability regions of the standard Gaussian distribution. 
For an $l$-bit symbol $i\in\{0,\ldots,2^l-1\}$, the corresponding interval is
\begin{equation}
    I_i=
    \left(
    \Phi^{-1}\!\left(\frac{i}{2^l}\right),
    \Phi^{-1}\!\left(\frac{i+1}{2^l}\right)
    \right],
\end{equation}
where $\Phi$ is the standard Gaussian CDF. 
A latent scalar is sampled from the Gaussian distribution truncated to the selected interval. 
Because the encrypted symbols are approximately uniform and the intervals have equal Gaussian probability, marginalizing over the symbols recovers the original standard Gaussian prior. 
Thus the watermark changes the latent region from which each scalar is sampled without changing the marginal distribution seen by the diffusion model.

Extraction reverses this procedure. 
An estimated terminal latent $\hat z_T$ is mapped back to its Gaussian regions, the corresponding symbols are decrypted, and repeated watermark bits are aggregated by voting. 
For the common setting $l=1$, the two regions are simply the negative and positive halves of the Gaussian distribution, so extraction reduces to recovering the sign of each latent coordinate. 
Therefore GS does not require exact reconstruction of the original terminal noise: it only requires the recovered latent to preserve sufficiently many of the correct Gaussian decision regions.
\section{Approach}

\subsection{One-Step Recovery as Conditional Regression}

Let $X_0$ denote the observed image-side latent, let $\epsilon$ denote the terminal Gaussian noise associated with the generated sample, and let $D$ denote an optional image perturbation for adversarial attack. We write a general one-step estimator as
\begin{equation}
    \hat\epsilon=f(\theta,X_0,D),
    \label{generalize_prediction}
\end{equation}
where $f$ denotes the complete estimator and $\theta$ denotes its learnable parameters.

Under squared endpoint regression,
\begin{equation}
    \min_\theta\;
    \E\!\left[
        \|f(\theta,X_0,D)-\epsilon\|_2^2
    \right],
    \label{eq:endpoint_regression}
\end{equation}
For Eq~\ref{eq:endpoint_regression}, the optimal output is the corresponding mean of noise conditional on $X_0$. Thus, the estimator is not required to identify the noise that generates the image; it only needs to predict the conditional endpoint statistic supported by the observation. Using $L_1$ regression, the optimization target is a conditional median.  Predicting only the mean or median doesn't matter, since if one image maps to multiple noises, then the DDIM inverse will sample a noise that is different from the original noise, but that almost never invalidates the GS. So we assume predicting the mean or median of noise in a limited region also still does not invalidate the GS, and use experiments to test it: training a neural network that does not aim to recover the endpoint, but returns the statistics of samples in a special region, to test whether it can properly extract the GS watermark.

For Gaussian Shading, this statistical relaxation is sufficient whenever the predicted endpoint remains in the same watermark decision region. Let $\operatorname{GSDec}_K$ denote the complete GS decoder. Our requirement is
\begin{equation}
    \operatorname{GSDec}_K(\hat\epsilon)
    =
    \operatorname{GSDec}_K(\epsilon),
    \label{eq:decision_statistic}
\end{equation}
rather than $\hat\epsilon=\epsilon$. This distinction is especially natural for the common $l=1$ setting, where each scalar is decoded primarily through its sign and repeated coordinates are aggregated by voting.

\subsection{High-SNR Endpoint Regression}

The conditional-regression view follows directly from the extended flow-matching formulation in Sec.~2. For an arbitrary coupled image--noise pair, let
\begin{equation}
    X_t=a_tX_0+b_t\epsilon
\end{equation}
define a non-causal interpolation with velocity $U_t$ used in the training. The corresponding causal field is
\begin{equation}
    v^*(x,t)=\E[U_t\mid X_t=x]=\frac{\dot{a}_t}{a_t}x+ (\dot{b}_t-\frac{\dot{a}_tb_t}{a_t})\E[\epsilon|X_t=x].
    \label{eq:conditional_velocity_ddpm_styl}
\end{equation}
Previous research \cite{liu2022rectified} already shows that, under the usual regularity assumptions, this causal velocity field induces the same marginal evolution as the non-causal interpolation. Importantly, the result does not require $X_0$ and $\epsilon$ to be independent.  When we use the original diffusion model to generate the noise-image pair, studying the means of the velocity will provide a different trajectory, where the velocity in the high-SNR area will point to the means of
possible noise that can generate this image.

Meanwhile, near the image endpoint, the input $X_t$ is very close to the final result, and predicting the velocity to finish the remaining trajectory is easy. So the watermark extractor that doesn't aim to reconstruct the complete noise-to-image/image-to-noise trajectory to reduce the computation cost can have a natural and simple training target: learn the endpoint statistic mean, since the original RF training target, as Eq. \ref{eq:conditional_velocity_ddpm_styl} shows, is equivalent to predicting the mean of possible endpoint $\epsilon$, and from the perspective of RF2, this is a simpler objective target than exact trajectory recovery. This observation motivates us to use a lightweight network rather than repeatedly evaluating a diffusion-scale denoiser. We can also further assume that predicting the median is feasible even when the training neural network is small.  We choose L1 loss to predict the median because previous research shows that DDPM \cite{ho2020ddpm}, which trains a denoiser and denoises noisy lantent/image step by step, is equivalent to the extended rectified flow\cite{liu2022rectified}, and the L1 loss in the denoising task has also been reported to provide better performance\cite{zhao2016loss}.
We call this model, which aims to recover the statistics of the sample in a limited region, the region-aware inversion network.

\subsection{Endpoint Parameterization and Perturbation}

Existing one-step estimators can also be written within Eq.~\eqref{generalize_prediction}. As one example, FARI uses the DDPM-style parameterization
\begin{equation}
    \hat\epsilon
    =
    a_T \widetilde X_0
    +
    b_T\,\epsilon_\theta(\widetilde X_0,0),
    \label{eq:fari_output}
\end{equation}
where
\begin{equation}
    a_T=\sqrt{\bar\alpha_T},
    \qquad
    b_T=\sqrt{1-\bar\alpha_T},
\end{equation}
and $X_0$ correspond to the latent in the image side, $\widetilde X_0=D(X_0)$. 



Another common setting of $\hat\epsilon$ is to directly predict the noise.  Explicitly, it's:
\begin{equation}
    \hat\epsilon = \epsilon_\theta(D(X_0))
    \label{simpleForm}
\end{equation}
We find that both of these two functions can successfully extract the watermark even if we use a very small neural network to optimize the regression loss.

Eq.~\ref{eq:fari_output} is based on DDPM, and we find that we can utilize the GPU more efficiently when we use a heuristic formula based on DDIM, which will be discussed in the next subsection. 


\subsection{Anchor--Residual Reverse Estimator}

Equation~\eqref{generalize_prediction} allows the complete estimator to contain multiple learned components, and Equation~\eqref{eq:fari_output} is just one of the special cases given by DDPM; we also provide a DDIM version. Our design is motivated by a simple optimization consideration: directly mapping an image-like latent concentrated near the low-dimensional manifold to a high-dimensional Gaussian noise vector requires a map between two very different types of representation. We therefore decompose the prediction so that the first network predicts an image-like target while the second network models the remaining noisy component. By decomposing the network into two parts, we can further utilize the GPU and augment parallel capabilities.

For each noisy image $X_t$, the original diffusion model $G_\theta$ can predict a clean image-like latent $X_0^*(t)$.
The first prompt-free network predicts an image-like anchor,
\begin{equation}
    \widehat X_0
    =
    F_0(D(X_0)).
    \label{eq:anchor_new}
\end{equation}
The output $\widehat X_0$ is a coarse clean representation, and its target is obtained from the original inference process.  Explicitly, in each timestep $t$, the diffusion model can input a noisy image $X_t$ and output a clean image $X_0^*(t)$. The training loss of $F_0$ is
\begin{equation}
    \mathcal L_1
    =
    \E\!\left[
        \operatorname{Dist}
        \bigl(F_0(D(X_0)),X_0^*(t)\bigr)
    \right],
    \label{eq:anchor_loss}
\end{equation}, and the $Dist$ can be $L2$ distance or $L1$ distance. In the training, we fix $t=0.9$, and $Dist$ to be $L1$ distance.

The second network predicts an intermediate representation conditioned on the observed latent,
\begin{equation}
    \widehat X_t
    =
    F_1\!\left(D(X_0)\right),
    \label{eq:intermediate_new}
\end{equation}
 A lightweight head predicts its effective noise level,
\begin{equation}
    \widehat t=F_2(\widehat X_t).
    \label{eq:timestep_head}
\end{equation}
Using the DDIM parameterization, the terminal noise estimate is reconstructed algebraically as
\begin{equation}
\begin{aligned}
    \hat\epsilon
    = f(\theta,X_0,D) 
    = \sqrt{1-\bar \alpha_T}
    \frac{
        \widehat X_t
        -
        \sqrt{\bar\alpha_{\widehat t}}\,
        SG(\widehat X_0)
    }{
        \sqrt{1-\bar\alpha_{\widehat t}}
    }
    + \sqrt{\bar \alpha_T}\,SG(\widehat X_0).
\end{aligned}
\label{eq:ddim_f}
\end{equation}
where $SG(\cdot)$ denotes stop gradient.

The second recovery loss directly supervises the final endpoint:
\begin{equation}
    \mathcal L_2
    =
    \E\!\left[
        \operatorname{Dist}
        \bigl(\hat\epsilon,\epsilon\bigr)
    \right].
    \label{eq:terminal_loss}
\end{equation}
Rather than fixing the intermediate noise level given by $t$, we allow the model to learn an effective $\widehat t$. During training, the timestep head detecting the noisy level of the output from $F_1$ is supervised on constructed interpolation states:
\begin{equation}
    X_t^*
    =
    \sqrt{\bar \alpha_t}\,X_0^*(0.9)
    +
    \sqrt{1-\bar \alpha_t}\,\epsilon,
    \label{eq:timestep_target}
\end{equation}
with
\begin{equation}
    \mathcal L_3
    =
    \E\!\left[
        \operatorname{Dist}
        \bigl(F_2(X_t^*),t\bigr)
    \right].
    \label{eq:timestep_loss}
\end{equation}

The complete training objective is
\begin{equation}
    \mathcal L_{\rm rev}
    =
    \lambda_0\mathcal L_1
    +
    \lambda_T\mathcal L_2
    +
    \lambda_{\rm reg}\mathcal L_3.
    \label{eq:our_loss_new}
\end{equation}

By utilizing this special structure, $F_0$ and $F_1$ can be inferred in parallel, and the GPU can be utilized more efficiently.  But it requires a small neural network to predict the \textit{noisy level of the $\widehat X_t$,} the timestep head introduces only a small additional cost.

Our implementation uses standard NAFNet~\citep{chen2022nafnet} backbones. No text encoder, prompt embedding, or cross-attention is required during extraction, and the source diffusion U-Net is absent from the learned reverse path.

\begin{figure*}[t]
    \centering
    \includegraphics[width=\textwidth]{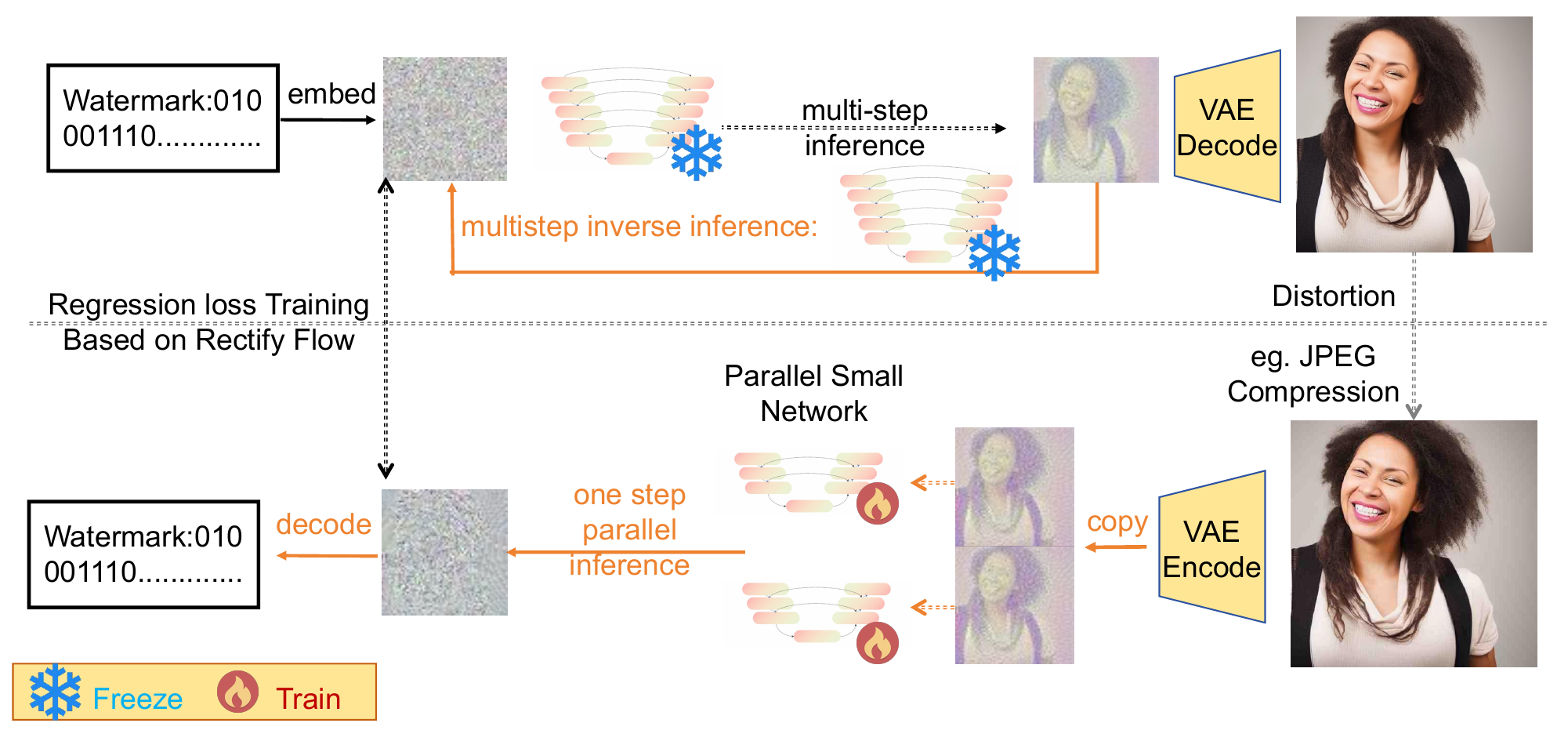}
    \caption{Extraction pipeline used in our implementation.}
    \label{fig:extract-algorithm}
\end{figure*}

\paragraph{Adversarial distortion training.}
Robustness is handled through the perturbation argument $D$ in Eq.~\eqref{generalize_prediction}. For each training sample, we construct a candidate set $\mathcal D_K$ containing common image corruptions such as JPEG compression, resizing, cropping, masking, blur, noise, and brightness changes. We select the corruption that maximizes the current reverse-recovery loss,
\begin{equation}
    d^*
    =
    \arg\max_{d\in\mathcal D_K}
    \mathcal L_{\rm rev}
    \bigl(f(\theta,X_0,d)\bigr),
    \label{eq:attack_new}
\end{equation}
and update the estimator using the selected example. This finite-set min--max procedure improves robustness without changing the underlying endpoint objective.  The complete pipeline is shown in Fig. \ref{fig:extract-algorithm}.

\section{Experiments}\label{sec:experiments}
\subsection{Ablation experiment}
\subsubsection{Current experimental setting}
To train our algorithm, the training set contains 89600 image-noise pairs generated offline.  During training, there is no additional cost to involve the original diffusion model; all necessary data is already generated and stored in a reverse-distillation LMDB feature store. 
We train six extractors to do the ablation study, and all of them are prompt-free and use width-64 NAFNet~\citep{chen2022nafnet} backbones with coordinatewise $L_1$ endpoint supervision.

The principal model is a two-branch reverse estimator given by Section 3.4.  Its $F_0$ branch and $F_1$ branch are both standard NAFNet\cite{chen2022nafnet}, each use two encoder stages with $[5,5]$ blocks, 12 middle blocks, and two decoder stages with $[5,5]$ blocks.  The second branch additionally predicts an effective timestep and reconstructs terminal noise using Eq.~\eqref{eq:ddim_f}.  We compare it with two single-network baselines, each using the larger $[10,10]$/24/$[10,10]$ NAFNet architecture: \textsc{Direct-$x_T$} directly regresses terminal noise, while \textsc{FARI like} predicts epsilon and converts it to $x_T$ with the terminal DDPM coefficients.  The objective target of the first network is described in Eq.~\ref{simpleForm}, and the objective target of the second network is described by Eq.~\ref{eq:fari_output}. All reported learned-model checkpoints correspond to 15,000 training iterations; the learning rate of the training is 1e-3; the optimizer is AdamW with a cosine annealing LR scheduler.  The training set contains 89600 image-noise pairs.  Meanwhile, all of the loss weights in Eq. \ref{eq:our_loss_new} are 1.0. The captions used to generate the offline image-noise training pairs are sampled from LAION \cite{schuhmann2022laion}.

To evaluate watermarks extraction, the text-to-image generation at $512\times512$ resolution use Stable Diffusion v1.5~\citep{rombach2022ldm} and COCO captions~\citep{lin2014coco}. Gaussian Shading's setting uses $f_c=1$, $f_{hw}=8$, $l=1$, assume $10^6$ users, and a target false-positive rate is $10^{-6}$.  Images are generated with the Diffusers UniPC scheduler \cite{zhao2023unipc} for 15 steps at guidance scale 5.5.  For every caption, we use a watermarked noise to generate an image; 
we extract the watermark via: 1. choose Gaussian noise and generate image, 2. $\beta$-VAE \cite{burgess2018understandingBeta} decode latent, 3. $\beta$-VAE encode image, 4. One-step extraction without caption.



\subsubsection{Metrics}

We follow the official Gaussian Shading~\cite{yang2024gaussianshading} evaluation protocol and report bit accuracy, detection rate, and traceability rate. For $l=1$, the recovered latent is converted to binary regions according to its sign and decoded using the official GS procedure. Detection and traceability are determined by the official thresholds $\tau_{\mathrm{onebit}}$ and $\tau_{\mathrm{bits}}$, respectively. We use $f_c=1$, $f_{hw}=8$, a target false-positive rate of $10^{-6}$, and $10^6$ candidate users.

\subsubsection{Ablation results}
We evaluate all six combinations on the same 1,000 cached COCO samples: direct $X_T$ prediction, Fari-Like Prediction, and DDIM heuristic prediction, with or without adversarial training. For every condition, all extractors receive the same attacked image and VAE-encoded latent. Table~\ref{tab:main_ab} shows that adversarial training consistently improves robustness under severe distortions while preserving clean extraction accuracy. The direct-$x_T$ and epsilon parameterizations remain closely matched, whereas the two-branch model is more sensitive without distortion training and recovers most of the gap after adversarial training.  

\begin{table*}[t]
\centering
\caption{SD~v1.5 extractor ablation under the matched ten-condition
image-distortion protocol on 1,000 COCO captions. All six
15,000-iteration checkpoints use the same cached images, GS states,
VAE pathway, and deterministic attack realizations.}
\label{tab:main_ab}

\scriptsize
\renewcommand{\arraystretch}{1.18}

\newcommand{\hdrcell}[1]{\shortstack[c]{#1}}

\newcommand{\paneltitle}[1]{%
    \addlinespace[2pt]
    \multicolumn{11}{c}{\textbf{#1}}\\[-1pt]
    \cmidrule(lr){1-11}
}

\newcommand{\tableheader}{%
    \textbf{Model}
    & \textbf{Clean}
    & \textbf{JPEG}
    & \hdrcell{\textbf{Crop/}\\\textbf{mask}}
    & \textbf{Drop}
    & \textbf{Resize}
    & \hdrcell{\textbf{G.}\\\textbf{blur}}
    & \textbf{Median}
    & \hdrcell{\textbf{G.}\\\textbf{noise}}
    & \textbf{S\&P}
    & \textbf{Bright.}\\
    \midrule
}

\resizebox{\textwidth}{!}{%
\begin{tabular}{@{}l@{\hspace{1.35em}}*{10}{c}@{}}

\toprule

\paneltitle{Bit accuracy (\%) $\uparrow$}
\tableheader

No-Adv Fari-Like
& 99.99\% & 97.70\% & 93.67\% & 93.33\% & 98.50\%
& 94.65\% & 99.04\% & 98.39\% & 91.31\% & 95.39\% \\

No-Adv Direct-$X_T$
& 99.99\% & 97.73\% & 93.70\% & 93.25\% & 98.51\%
& 94.63\% & 99.06\% & 98.41\% & 91.33\% & 95.34\% \\

No-Adv Two brach
& 99.99\% & 97.63\% & 91.59\% & 90.82\% & 98.42\%
& 94.38\% & 98.99\% & 98.38\% & 91.28\% & 93.02\% \\

\addlinespace[2pt]

Adv Fari-Like
& 100.00\% & 98.31\% & 94.51\% & 94.50\% & 99.03\%
& 97.65\% & 99.26\% & 98.78\% & 93.81\% & 96.43\% \\

Adv Direct-$X_T$
& 99.99\% & 98.32\% & 94.51\% & 94.49\% & 99.04\%
& 97.68\% & 99.26\% & 98.79\% & 93.87\% & 96.47\% \\

Adv Two brach
& 99.99\% & 98.15\% & 93.72\% & 93.64\% & 98.93\%
& 97.24\% & 99.17\% & 98.67\% & 93.12\% & 96.09\% \\

\toprule
\paneltitle{Detection rate (\%) $\uparrow$}
\tableheader

No-Adv Fari-Like
& 100.00\% & 99.70\% & 100.00\% & 100.00\% & 99.90\%
& 100.00\% & 100.00\% & 99.90\% & 98.70\% & 97.60\% \\

No-Adv Direct-$X_T$
& 100.00\% & 99.70\% & 100.00\% & 100.00\% & 100.00\%
& 100.00\% & 100.00\% & 99.90\% & 98.80\% & 97.50\% \\

No-Adv Two brach
& 100.00\% & 99.60\% & 100.00\% & 100.00\% & 100.00\%
& 99.90\% & 100.00\% & 99.90\% & 98.50\% & 95.10\% \\

\addlinespace[2pt]

AdvFari-Like
& 100.00\% & 99.80\% & 100.00\% & 100.00\% & 100.00\%
& 100.00\% & 100.00\% & 99.90\% & 99.20\% & 97.80\% \\

Adv Direct-$X_T$
& 100.00\% & 99.80\% & 100.00\% & 100.00\% & 100.00\%
& 100.00\% & 100.00\% & 99.90\% & 99.40\% & 98.10\% \\

Adv Two brach
& 100.00\% & 99.70\% & 100.00\% & 100.00\% & 100.00\%
& 100.00\% & 100.00\% & 99.90\% & 99.20\% & 97.80\% \\

\toprule
\paneltitle{Traceability rate (\%) $\uparrow$}
\tableheader

No-Adv Fari-Like
& 100.00\% & 99.40\% & 100.00\% & 100.00\% & 99.70\%
& 99.70\% & 100.00\% & 99.50\% & 95.70\% & 95.80\% \\

No-Adv Direct-$X_T$
& 100.00\% & 99.40\% & 100.00\% & 100.00\% & 99.80\%
& 99.70\% & 100.00\% & 99.60\% & 95.80\% & 95.90\% \\

No-Adv Two brach
& 100.00\% & 99.20\% & 100.00\% & 99.90\% & 99.70\%
& 99.60\% & 100.00\% & 99.70\% & 96.10\% & 92.00\% \\

\addlinespace[2pt]

Adv Fari-Like
& 100.00\% & 99.40\% & 100.00\% & 100.00\% & 99.80\%
& 99.80\% & 100.00\% & 99.70\% & 98.60\% & 97.10\% \\

Adv Direct-$X_T$
& 100.00\% & 99.50\% & 100.00\% & 100.00\% & 99.90\%
& 99.90\% & 100.00\% & 99.80\% & 98.30\% & 97.40\% \\

Adv Two brach
& 100.00\% & 99.10\% & 100.00\% & 100.00\% & 99.80\%
& 99.70\% & 100.00\% & 99.70\% & 97.90\% & 96.50\% \\

\bottomrule
\end{tabular}}
\end{table*}

Table~\ref{tab:nafnet_latency} reports extractor-only wall-clock measurements on 100 clean cached images. This experiment deploys the neural networks on two RTX PRO 6000 GPUs. We use FP32 inference, batch size 10, and three warm-up batches on the same two-GPU system. The single-NAFNet models use data parallelism, splitting each batch equally across the two devices, while the two-branch model places the $x_0$ and $x_T$ branches on separate devices. Host-to-device input transfer, VAE encoding, and GS decoding are excluded; cross-device output gathering and terminal-noise reconstruction are included. The two-branch allocation requires $2.993$--$3.085$ ms per image, compared with $4.847$--$4.977$ ms for the single-network alternatives. Adversarial training changes the weights but not the architecture and has little effect on extraction latency.

\begin{table*}[t]
\centering
\begin{minipage}{\columnwidth}
\centering
\caption{Latency for extracting 100 watermarks with the extractor deployed on two RTX PRO 6000 GPUs. Single-network models use data parallelism; the two-branch model uses branch parallelism. Lower latency and higher throughput are better.}
\label{tab:nafnet_latency}
\small
\renewcommand{\arraystretch}{1.10}
\setlength{\tabcolsep}{4pt}
\begin{tabularx}{\columnwidth}{@{}Xrrrrrr@{}}
\toprule
Method & Total (ms) $\downarrow$ & ms/image $\downarrow$ & images/s $\uparrow$ & GFLOPs & GMACs & params \\
\midrule
No-adv Direct $x_T$ & 489.258 & 4.893 & 204.39  & \textbf{13.833} & \textbf{6.873} & \textbf{14,539,460}\\
No-adv Fari-like & 484.709 & 4.847 & 206.31  & \textbf{13.833} & \textbf{6.873} & \textbf{14,539,460}\\
No-adv Two-branch & \textbf{299.330} & \textbf{2.993} & \textbf{334.08} & 14.394 & 7.148  &15,467,792   \\
\addlinespace
adv Direct $x_T$ & 485.419 & 4.854 & 206.01 & \textbf{13.833} & \textbf{6.873} & \textbf{14,539,460}\\
adv Fari-like & 497.652 & 4.977 & 200.94  & \textbf{13.833} & \textbf{6.873} & \textbf{14,539,460}\\
adv Two-branch & \textbf{308.505} & \textbf{3.085} & \textbf{324.14} & 14.394 & 7.148  &15,467,792   \\
\bottomrule
\end{tabularx}
\end{minipage}
\end{table*}

\subsection{Comparison with SOTA extraction}
\subsubsection{Robustness comparison protocol}
For the matched SOTA comparison, we use Stable Diffusion v2.1-base for both generation and extraction. We evaluate the DDIM inverse 50-step, the official FARI, the official OSI checkpoint and our adversarially trained SD~v2.1 two-branch checkpoint on the same 1,000 COCO captions, seeds, GS keys, $512\times512$ images, 15-step UniPC schedule, and VAE re-encoding pathway. Each generated image is independently evaluated under all ten distortions given by FARI\cite{faricode2026}: the unmodified VAE decode-encode roundtrip; JPEG quality 25; crop/mask ratio 0.6; random-drop ratio 0.8; resize ratio 0.25; Gaussian blur radius 4; median-filter kernel 7; additive Gaussian noise with standard deviation 0.05 on the $[0,1]$ image scale; salt-and-pepper probability 0.05; and brightness factor 6. Attack randomness is deterministically seeded per caption and condition. Table \ref{tab:robustness} reports mean GS bit accuracy, detection rate, and traceability rate.  Meanwhile, to ensure the algorithm doesn't memorize, we also use random Gaussian noise that hasn't been augmented with GS to generate images, and extract the corresponding noise, report the Max bit accuracy given by comparing those extracted noise with all cached noise with watermark. 
\begin{table*}[t]
\centering
\caption{Matched SD~v2.1 robustness comparison under the FARI-style image-distortion protocol on 1,000 COCO captions. The first group reports bit accuracy on watermarked images. The second group reports the \emph{maximum} bit accuracy when we compare the cached watermark with the extracted bit strings given by the \emph{controlled} images group generated from the same prompts but with different noise and \emph{no watermark}.}
\label{tab:robustness}
\small
\setlength{\tabcolsep}{3pt}
\resizebox{\textwidth}{!}{%
\begin{tabular}{lcccccccc}
\toprule
& \multicolumn{4}{c}{Mean bit accuracy (watermarked) $\uparrow$}
& \multicolumn{4}{c}{Max bit accuracy (controlled) $\uparrow$} \\
\cmidrule(lr){2-5}\cmidrule(lr){6-9}
Distortion & FARI & OSI & Ours & DDIM inv (50) & FARI & OSI & Ours & DDIM inv (50) \\
\midrule
Clean VAE roundtrip & 99.99\% & 100.00\% & 99.99\% & 100.00\% & 64.45\% & 65.63\% & 64.84\% & 64.06\% \\
JPEG (Q=25) & 98.33\% & 99.09\% & 97.81\% & 97.63\% & 64.45\% & 64.84\% & 65.62\% & 64.45\% \\
Random crop/mask (0.6) & 95.63\% & 97.67\% & 93.65\% & 95.52\% & 64.84\% & 64.45\% & 64.06\% & 64.84\% \\
Random drop (0.8) & 95.09\% & 97.56\% & 93.49\% & 94.93\% & 64.45\% & 64.06\% & 64.84\% & 64.06\% \\
Resize (0.25) & 99.09\% & 99.48\% & 98.65\% & 98.59\% & 65.23\% & 64.84\% & 65.23\% & 64.45\%\\
Gaussian blur (r=4) & 97.55\% & 98.99\% & 96.61\% & 95.18\% & 64.84\% & 64.84\% & 65.23\% & 65.63\% \\
Median blur (k=7) & 99.22\% & 99.67\% & 98.93\% & 99.03\% & 64.84\% & 64.06\% & 64.45\% & 65.63\% \\
Gaussian noise ($\sigma=0.05$) & 98.79\% & 99.52\% & 98.47\% & 98.62\% & 64.84\% & 64.45\% & 64.84\% &65.63\% \\
Salt-and-pepper (p=0.05) & 94.14\% & 99.99\% & 92.42\% & 91.17\% & 65.63\% & 66.02\% & 63.67\% & 66.41\% \\
Brightness (factor=6) & 95.97\% & 96.62\% & 95.48\% & 95.20\% & 65.63\% & 64.45\% & 64.06\% & 64.45\% \\
\bottomrule
\end{tabular}}
\end{table*}

Table~\ref{tab:extraction_cost} compares the backbone cost of one latent-to-noise watermark-extraction pass. Our lightweight two-branch extractor requires only $14.394$ GFLOPs and $7.148$ GMACs, reducing computation by approximately $47.4\times$ relative to FARI and $47.2\times$ relative to OSI in FLOPs. It also uses only $15.468$M total parameters, compared with $868.469$M for FARI and $865.911$M for OSI. FARI trains only $2.558$M LoRA parameters, but extraction still executes and stores the frozen SD~v2.1 U-Net. For a matched backbone comparison, image generation, latent encoding (including OSI's encoder and quantizer), and GS decoding are excluded. We also report the latency of watermark extraction on 224 images using 2 A800 GPUs, with FARI and OSI's batch size set to 2 while our method's batch size is 112.  By using this batch size, the peak GPU allocation of FARI is 3518/3510 MiB, the peak GPU allocation of OSI is 3498/3489 MiB, and the peak GPU allocation of ours is 1098/1816 MiB. The extraction latency of FARI and OSI is 37.34 and 31.32 ms/image, while our is 1.39 ms/image.

\begin{table*}[t]
\centering
\begin{minipage}{\columnwidth}
\centering
\caption{Backbone cost of one latent-to-noise watermark extraction, measured with \texttt{calflops}; lower is better. Image encoding and GS decoding are excluded for all methods.}
\label{tab:extraction_cost}
\small
\setlength{\tabcolsep}{4pt}
\begin{tabular}{@{}lrrrr@{}}
\toprule
Method & GFLOPs $\downarrow$ & GMACs $\downarrow$ & Params (M) $\downarrow$ & latency (ms/image)$\downarrow$ \\
\midrule
FARI & 682.675 & 341.079 & 868.469 & 37.344 \\
OSI & 678.723 & 339.103 & 865.911 & 31.320 \\
\textbf{Ours} & \textbf{14.394} & \textbf{7.148} & \textbf{15.468} &\textbf{1.392} \\
\bottomrule
\end{tabular}
\end{minipage}
\end{table*}


We also report the number of false positives in Table \ref{tab:fpr}. Simultaneously, we report the watermark diversity there to ensure a fair comparison: the number of keys, nonces, and messages we use in the GS is 1000, and all of them are unique. And the random seed to initiated training is different from each seed use to get the initiate Gaussian noise.

\begin{table*}[t]
\centering
\caption{The number of false positive on \emph{controlled} images with \emph{no watermark}. For each distortion, we run extraction, recover a bit string, and declare a detection if it passes a threshold after comparing it with cached watermarks. We do 1,000,000 comparisons and, based on the max bit accuracy, determine the detection result;}
\label{tab:fpr}
\scriptsize
\setlength{\tabcolsep}{3pt}
\renewcommand{\arraystretch}{1.12}
\resizebox{\textwidth}{!}{%
\begin{tabular}{lcccccccccc}
\toprule
\textbf{Extractor} & \textbf{Clean} & \textbf{JPEG} & \textbf{Crop/mask} & \textbf{Drop} & \textbf{Resize} & \textbf{GBlur} & \textbf{MBlur} & \textbf{GNoise} & \textbf{S\&P} & \textbf{Bright} \\
\midrule
FARI & 0 & 0 & 1 & 0 & 3 & 1 & 1 & 3 & 1 & 1 \\
OSI & 3 & 2 & 0 & 0 & 1 & 2 & 0 & 0 & 2 & 0 \\
Ours & 1 & 3 & 0 & 1 & 1 & 1 & 0 & 1 & 0 & 0 \\
DDIM inv (50) & 0 & 0 & 1 & 0 & 0 & 2 & 1 & 1 & 2 & 0 \\
\bottomrule
\end{tabular}}
\end{table*}

\section{Conclusion}
Gaussian Shading does not require a detector to reconstruct a unique floating-point terminal latent.  Its cryptographic sampler partitions the Gaussian prior into finitely many equal-probability regions, and the final decoder only requires enough of those region labels to survive decryption and repetition voting.  Extended flow matching and DDPM-style $\epsilon$ regression make the complementary statistical point: regression from a partially observed state naturally learns conditional statistics of hidden endpoints, not sample-specific hidden pairs.  These two facts together motivate direct one-step watermark extraction.

We therefore formulate GS extraction as decision-region recovery and propose a prompt-free two-NAFNet reverse estimator trained with intentional $L_1$ loss and finite-set adversarial image distortions.  The two-branch anchor/interpolation parameterization is an optimization method rather than a claim that a new continuous trajectory has been derived.  

\section{Further Discussion}
\paragraph{Revisiting the trajectory interpretation of one-step extraction.}
FARI uses a simple regression loss, which is different from the SOTA distillation loss~\cite{yin2024dmd,yin2024improveddmd,luo2311lcm,song2023improvedlcm} that tends to generate a sample from the target distribution directly in one step.  The objective target given by FARI is the one equivalent to the extended RF2 algorithm in the high SNR stage~\cite {liu2022rectified}.  And whether it can generate a sample in one step is determined not by the trajectory characteristics of the extended RF-1 algorithm, but by the trajectory characteristics of the extended RF-2 algorithm.  FARI's trajectory interpretation effectively treats watermark extraction as a sample-generation problem, but this interpretation is questionable.


\paragraph{Latest DMD2 algorithm remove regression loss for the sake of the image generation.}
Another evidence that the regression loss would not let the model directly generate a real sample from the target distribution is the DMD2 \cite{yin2024improveddmd} algorithm which remove the regression loss since it violate the distribution matching target.
\bibliography{references}
\end{document}